\documentclass{article}
\usepackage{spconf,amsmath,graphicx}

\title{Segmentation-by-Detection: A Cascade Network for Volumetric Medical Image Segmentation}
\name{Min Tang, Zichen Zhang, Dana Cobzas, Martin Jagersand, Jacob L. Jaremko}
\address{
    University of Alberta\\
	Edmonton, Canada\\
	}
	
\begin{document}
%
\maketitle
\begin{abstract}
We propose an attention mechanism for 3D medical image segmentation. The method, named segmentation-by-detection, is a cascade of a detection module followed by a segmentation module. The detection module enables a region of interest to come to attention and produces a set of object region candidates which are further used as an attention model.
Rather than dealing with the entire volume, the segmentation module distills the information from the potential region. This scheme is an efficient solution for volumetric data as it reduces the influence of the surrounding noise which is especially important for medical data with low signal-to-noise ratio. Experimental results on 3D ultrasound data of the femoral head shows superiority of the proposed method when compared with a standard fully convolutional network like the U-Net. 

\end{abstract}
\section{Introduction}
\label{sec:intro}

Segmentation is essential in medical image analysis to identify the structures of interest, such as organs or tumors.
Despite numerous approaches have been proposed over the past decade, it continues to be a challenging problem as medical images are complex in nature and rarely have any simple linear feature. 
Most traditional technique \cite{Boykov06GC,Caselles97geodesics,ChanVese01regions} utilize handcrafted features incorporated into an energy-based segmentation method or a machine learning classifier. Those methods can easily incorporate different types of regularizers and priors, making the segmentation problem easier and more flexible \cite{Cremers-et-al-ijcv07}. However, their performances rely on a good contour initialization, good parameter selection that is often done in a heuristic way, and the quality of the handcrafted features. Moreover, those traditional models are not flexible enough to account for the large appearance variation and abnormalities present in medical imaging. 

Deep convolutional neural networks (CNNs) have rapidly developed as a powerful tool in computer vision and pattern recognition and are recently widely used in anatomy delineation \cite{litjens2017survey}. The success of CNNs owe to the hierarchical representations automatically learned from the raw input rather than the traditional handcrafted features. 
Early CNNs formulate the segmentation problem as a pixel-wise classification by presenting local region around that pixel. This sliding window approach only considers local context and suffer from efficiency issues.
Fully Convolutional Network (FCN) \cite{long2015fully} addressed these issues by an end-to-end, pixels-to-pixels architecture which can be applied on an entire input image and prevent the decrease in resolution. Many deep architectures were proposed based on this model. One of the most successful network in medical image segmentation is the U-Net \cite{ronneberger2015u}. This architecture consists of a contracting path and a symmetric expanding path. The skip connections between these two paths enable global context as well as precise localization. 

Most image segmentation systems fine-tuned a pre-trained network to try to work around the requirement of large data sets for deep network training. 
However, medical data is often 3D (e.g. MRI or CT volume) while the majority of pre-trained CNNs are 2D. Therefore, a slice-by-slice analysis is often adopted to exploit the pre-trained nets \cite{prasoon2013deep}. In this way the computation is less expensive but an important drawback is that the inter-slice anatomical context is not explored.
The 3D architectures \cite{cciccek20163d,milletari2016v} take on a whole volume to take the contextual information between slices into account. Still, the high cost in computation and low speed of convergence in training process are issues that remain unsolved.

The success of region proposal methods, especially the Region Proposal Networks (RPN) \cite{ren2015faster} have demonstrated the effectiveness of the attention mechanism in object detection. 
Localization is an easier problem than segmentation. Thus, 
we can take the advantage of the attention mechanism in segmentation task and propose a cascade network for 3D volumetric data, named segmentation-by-detection.
Rather than focusing on an entire volume, the 2D detection module allows for region of interest (RoI) to come to attention as needed. This is especially important for medical data with low signal-to-noise ratio (SNR). Using representations that distill information in the potential region rather than a volume is also an efficient solution for volumetric data.

Intuitively, our framework consists of a 2D detection module based on RPN \cite{ren2015faster} and a 3D segmentation module that follows the U-Net architecture. Compared to other recent works on 3D FCNs \cite{milletari2016v,cciccek20163d,dolz20173d,he2017mask}, our contribution are two-fold. First, we propose a novel cascade architecture to combine the attention mechanism with the segmentation system, 
which allows for learning from potential object regions.
Second, the architecture can benefit from the pre-trained models by exploiting the learned weights in the detection module. Besides, the method used in each module is flexible to change depending on the data and segmentation task.
Experiments on 3D ultrasound data of femoral head shown that the proposed architecture is more accurate and efficient when compared to 3D U-Net and other related volumetric segmentation methods.

\section{Related Work}
\textbf{Region Proposal Methods:} 
Region proposal algorithms are widely used for object detection in natural images \cite{girshick2014rich}. Usually, a set of rectangular region proposals are generated as candidate object regions for further processing. The region proposal method allows for local RoI to come to attention and reduce the interference of the global context which is especially meaningful for medical data with low SNR. 

RPN is currently the leading method. It was introduced in Faster R-CNN \cite{ren2015faster} and extended by Mask R-CNN \cite{he2017mask} for object instance segmentation. 
In the context of segmentation, a precise object localization with a bounding box helps correct labeling. Mask R-CNN implements this idea by adding a branch for predicting the segmentation mask, in parallel with the branch for classification and bounding box regression. Different from the above, 
our network is a cascade of a detector followed by a segmentation module and targeted for 3D volume data.
\vspace{1em}

\noindent\textbf{Volumetric Segmentation:} 
Early CNN-based methods in medical image segmentation often use a slice-by-slice analysis. In order to exploit the anatomic context between slices, many 3D architectures have been proposed recently. 3D U-Net \cite{cciccek20163d} is an extension of the 2D architecture \cite{ronneberger2015u} which combines context information from lower layers with semantically richer features from the higher layers. The segmentation module of the proposed method is based on 3D U-Net and learned from densely annotated data instead of sparsely annotated ones. 

Likewise, V-Net \cite{milletari2016v} utilizes the encoder and decoder scheme and proposes a Dice coefficient based loss layer for segmentation task. In \cite{dolz20173d}, a 3D fully CNN was proposed for subcortical structure segmentation which uses small kernels and deeper architecture to avoid computation and memory burden.
The common idea of all these architectures \cite{cciccek20163d,milletari2016v,dolz20173d} is to parse an entire volume, which makes the system somewhat inefficient. The proposed segmentation-by-detection architecture address that by integrating the attention mechanism that guides the network to learn from the potential RoI rather than the whole target volume.

\section{Method}
\label{sec:method}

\begin{figure}[t]
\centering
\includegraphics[width=0.95\linewidth]{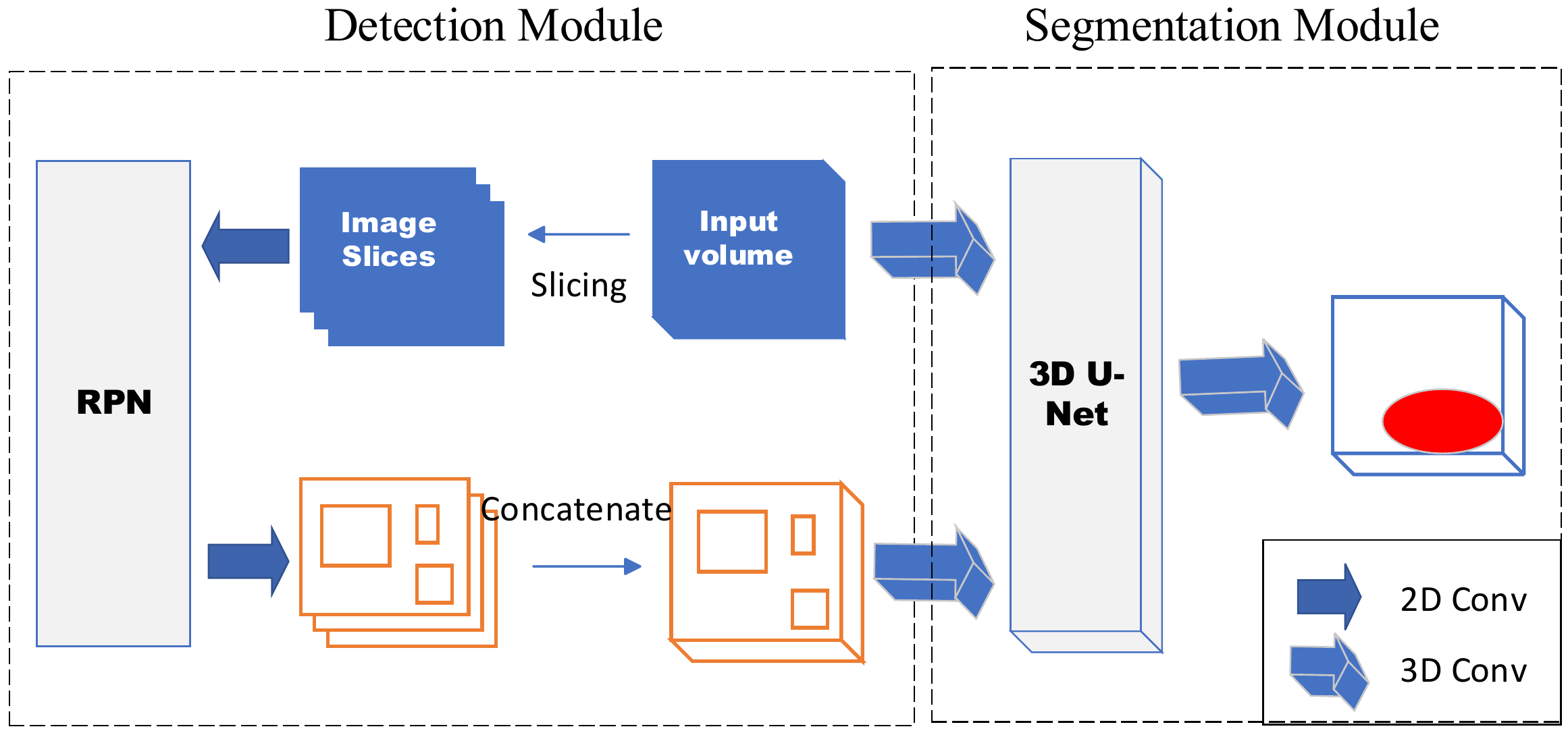}
\caption{Schematic representation of the segmentation-by-detection framework. The left part is the detection module while the segmentation module is followed on the right. The blue block denotes the input volume which is 3D ultrasound scan of femoral head. The output segmentation is in red.}
\label{fig:framework}
\end{figure}

The proposed segmentation-by-detection framework, as depicted in Figure \ref{fig:framework}, consists of a detection module and a segmentation module.
In detection stage, 2D slices (layered box) from the input volume are fed to the RPN. Based on the region proposals obtained from RPN, an attention model (block in orange) is formed. The input volume as well as the attention model are further processed in segmentation stage to get the refined anatomical segmentation. 
\vspace{1em} 

\noindent\textbf{Detection Module:} 
The detection module follows an RPN architecture, a fully convolutional network which takes image slice as input and outputs object region candidates. 
We use the VGG-16 model as the backbone \cite{simonyan2014very} to learn convolutional features and an $3 \times 3$ spatial window to generate region proposals. At each sliding-window location, 9 anchors are predicted associated with different scales and aspect ratios. The last layer consists of a box-regression (reg) layer and a box-classification (cls) layer in parallel. The reg layer outputs 4 regression offsets, $ t = (t_x,t_y,t_w,t_h)$, denoting a scale-invariant translation as well as log-space height and width shift, where $x,y,w$ and $h$ specify two coordinates of the box center, width and height. The cls layer outputs two scores by softmax, related to probabilities of object and background for each proposal. We assign a positive label (of being object) to candidate which has an Intersection-over-Union (IoU) ratio higher than 0.7 with ground truth box. Note that an image slice may contain multiple object regions or none. 

The loss function of RPN follows the multi-task loss \cite{ren2015faster} which is defined as $L = L_{reg} + L_{cls}$. The regression loss, $L_{reg} = -\log p_{obj}$ is log loss and the classification loss,
\begin{equation} \label{eq:loss}
L_{cls} = \sum_{i \in \{x,y,w,h\}} smooth_{L_1} (t_i - t_i^*)
\end{equation}
is smooth $L_1$ loss where $t_i^*$ denotes the ground truth box for the target object. 
\vspace{1em}

\noindent\textbf{Segmentation Module:}
3D U-Net \cite{cciccek20163d} is utilized in the segmentation module as its outstanding performance in medical image segmentation. The u-shaped architecture consists of two paths: a contracting path, where each layer contains two $3\times3\times3$ convolutions followed by a rectified linear unit (ReLU) and then a max pooling, provides high resolution features. While, the symmetric expanding path for semantically richer features replaces max pooling with a upconvolution $2\times2\times2$ with stride of 2 in each dimension, and then two $3\times3\times3$ convolutions each followed by a ReLU. Skip connections between layers of equal resolution in the contracting path and the expanding path enables context information as well as precise localization.

Different from 3D U-Net, to incorporate the attention model detected by the RPN, our architecture takes as input both the volumetric image data and the candidate RoIs proposed by the RPN, concatenated as 3D volume. 
The attention model makes the network to focus on the potential RoIs and can reduce the interference of the surrounding noise.
The anatomical segmentation is then generated from a $1\times1\times1$ convolution which reduces the number of feature maps to the number of labels.  The energy function is computed by a pixel-wise softmax combined with the cross entropy loss.

\subsection{System and implementation Details}
The segmentation-by-detection approach adopts a cascade structure with two stages: detection and segmentation. The two networks are trained separately in an end-to-end manner. All the new layers are randomly initialized from zero-mean Gaussian distribution with standard deviations 0.01. Biases are initialized to 0. We use Caffe \cite{jia2014caffe} for the implementation and an NVIDIA Titan X GPU for training.

In the detection stage, we initialize the VGG-16 model by the pre-trained model for ImageNet classification \cite{russakovsky2015imagenet} and further fine-tune the model for our detection task. The input fed to the network are image slices with a fixed size of $184\times96$ and the corresponding ground truth boxes are generated from the annotation in the format of tight bounding boxes surrounding the segmentation contour (as illustrated in Figure \ref{fig:hip} (b), the boundary of white area). To optimize the energy function, stochastic gradient descent (SGD) is used. The global learning rate is set to 0.001, while a momentum of 0.9 and a weight decay of 0.0005 are used. The batch size is set to 256 and each mini-batch only contains the positive anchors for training. The region proposals are obtained from the reg path for each image slice. The attention model is then formed by concatenating all the detected regions, as binary masks, into a volume.

In the segmentation stage, we use the Adam optimizer \cite{kingma2014adam} to learn the network parameters. A global learning rate is set to 0.001 while the two momentum coefficients are set to 0.9 and 0.999 respectively. A batch size of 1 is used due to the memory constraints of the GPU. The network takes the volume data as well as the attention model as input. We train the network for a maximum of 30K iterations and reserve the learned weights with the best performance from every 1K iterations. 
\vspace{1em}

\noindent\textbf{Inference:}
At test time, the 2D slices from an input volume are first fed to the detection module. The attention model is obtained based on the output. Then the volume data as well as the attention model are fed to the segmentation module to get the pixel-wise prediction.

\section{Experiments and Results}
\label{sec:exp}

\begin{figure}[tb]
\centering
\centerline{\includegraphics[width=.45\textwidth]{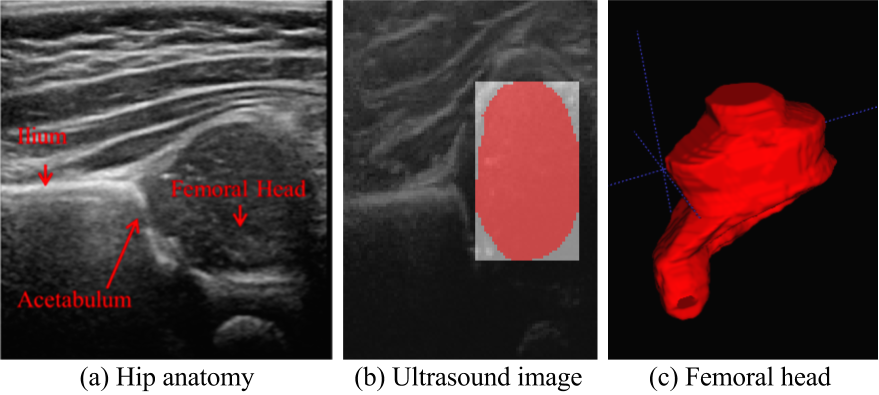}}
\caption{An illustration of ultrasound image for the hip anatomy. (a) Hip anatomy; (b) Ultrasound image in 2D view where the red area is annotation while the white area is ground truth object region; (c) Femoral head in 3D view.}
\label{fig:hip}
\end{figure}

\noindent
{\bf Data} 
The proposed method was evaluated on a segmentation task for 3D ultrasound images of infants, collected by a team at the Radiology Department at University of Alberta. The data is part of a local study of developmental dysplasia of the hip (DDH)\cite{hareendranathan2017semiautomatic}. It was shown that  with a correct early diagnosis, DDH can be effectively treated by simple bracing in infancy. This diagnosis is currently done using ultrasound images. The segmentation of the hip is an essential step of this procedure and therefore an important task that we address in this work. An illustration of ultrasound image for the hip anatomy is shown in Figure \ref{fig:hip}.


A total of 19 three-dimensional ultrasound (3DUS) scans from 19 infants aged between 4 and 111 days are used. 3DUS volumes of the hip are obtained on a Philips iU22 scanner using a 13MHz linear transducer in coronal orientation and exported to Cartesian DICOM. Each 3DUS consists of 256 ultrasound slices of 0.13mm thickness containing $397\times192$ pixels measuring $0.11\times0.20mm$. We ran all our experiments on a down-sampled version of the original resolution by factor of two in each dimension. 
Therefore, the data size used in the experiments is $128\times184\times96$. The training set contains 5 left femoral head and 5 right femoral head randomly selected from 3DUS scans, while the remaining scans are chosen for testing.

\begin{figure}[t!]
\centering
\centerline{\includegraphics[width=.46\textwidth]{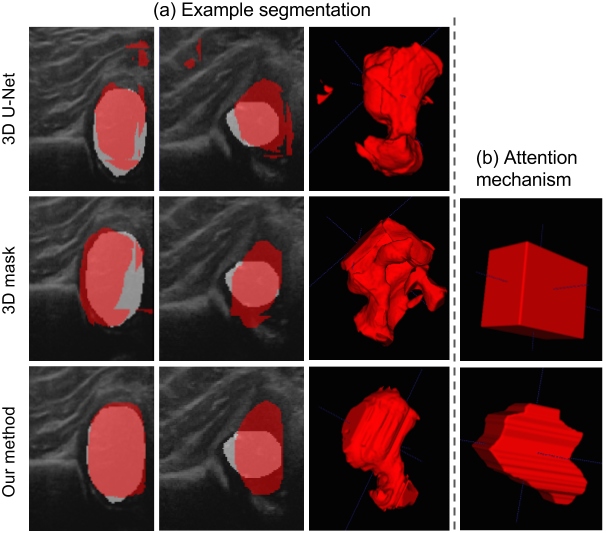}}
\caption{Segmentation results on femoral head. (a) Example segmentation. Output segmentation is in red while the ground truth is in white. The top and middle rows show the results of 3D U-Net and 3D mask respectively while our method is displayed at the bottom. The first two columns are the segmentation results in image slices. The third column displays the 3D view of segmented femoral head. (b) Attention mechanism. The above is the 2D attention model while the bottom is the 3D mask.}
\label{fig:seg}
\end{figure}

\noindent
{\bf Comparative experiments}
To evaluate the effect of the proposed method, we compared it with the 3D U-Net. 
Different from the training process of 3D U-Net, the proposed method is trained with the input volume as well as the attention model obtained from the detection module. 
In addition, to demonstrate the effectiveness of the proposed attention mechanism, we compared our 2D attention model with a 3D mask, which can be easily generated from the annotation in the format of a 3D tight bounding box as illustrated in Figure \ref{fig:seg} (b). The 3D mask method replaces the detection module with the mask while the segmentation module is unchanged. 

We evaluate the three segmentation methods using IoU, which is defined as $TP/(TP + FP + FN)$ where $T/P$ denotes $True/False$ and $P/N$ denotes $Positive/Negative$. The comparative models were tested on 9 scans separately and results were reported as the best performance, the worst as well as the average IoU. 

\begin{figure}[t!]
\centering
\centerline{\includegraphics[width=.5\textwidth]{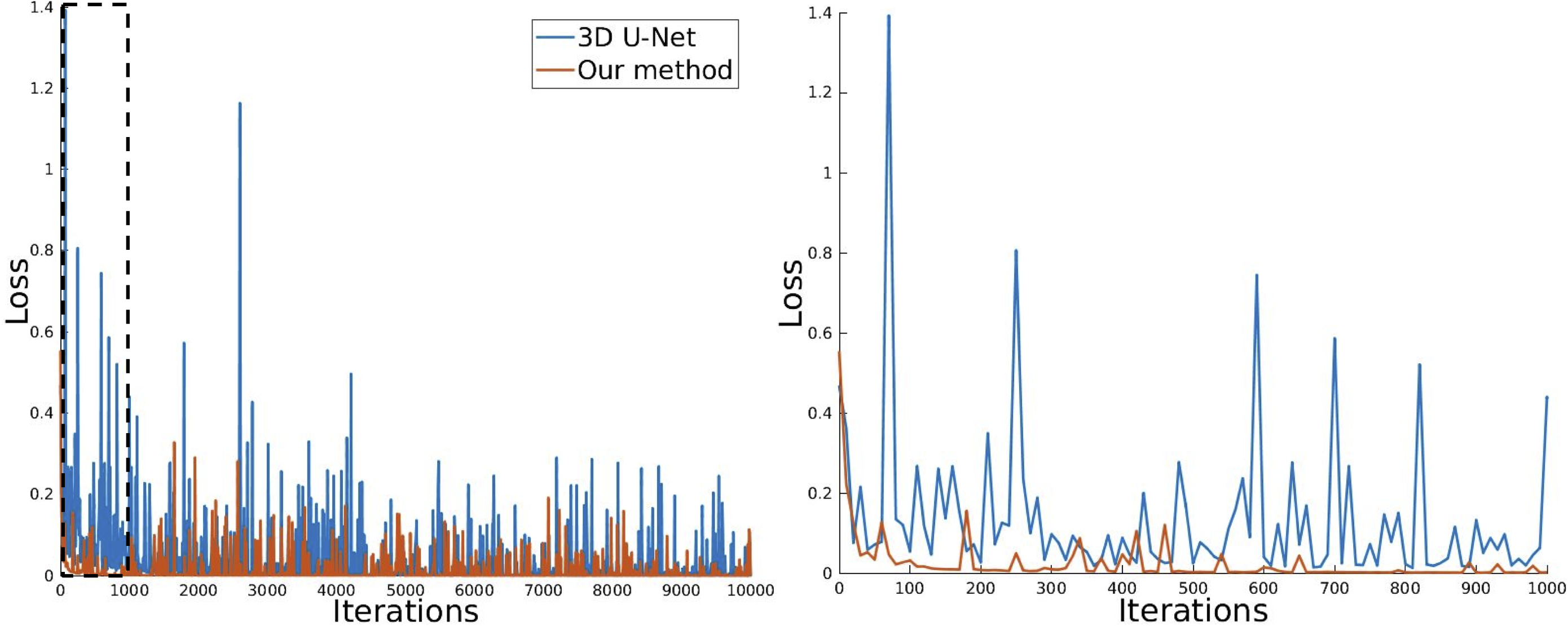}}
\caption{Training loss. The loss of 3D U-Net is denoted by the blue line while ours is denoted by red. The right is a zooming in of the dashed region.}
\label{fig:loss}
\end{figure}

\noindent
{\bf Results} 
Summary results are presented in Table \ref{tb:unet}.
Compared with the 3D U-Net, our method improves the overall performance (average IoU is increased by 8.7\%). The segmentation results are illustrated in Figure \ref{fig:seg} (a). With the guidance of attention model, our method reduced the disrupt of global context and shows a smoother segmented contour. Compared with the left column, results in the middle are low in precision. This is reasonable because of the low contrast in gray scale which is hard to annotate even for the expert. The training loss is also depicted in Figure \ref{fig:loss}, where our method showed a faster convergence speed and more stable training.

When comparing results with the 3D mask, our attention model showed an obvious advantage. It is mainly because the slice-by-slice formed attention model is more precise in localization than the 3D bounding box. As the performance of the detection module increases, the segmentation gets better. That means, we can improve the performance of segmentation by devoting efforts on detection which is an easier task.

\begin{table}[thb]
\centering
\caption{Segmentation results for femoral head}
\begin{tabular}{l|ccc}
\hline
Femoral head & Best IoU & Worst IoU & Average IoU \\ 
\hline
3D U-Net     & 0.729    & 0.483     & 0.622 \\
3D mask & 0.679    & 0.493     & 0.600 \\
Our method   & {\bf 0.779}    & {\bf 0.512}     & {\bf 0.709} \\
\hline
\end{tabular}
\label{tb:unet}
\end{table}

\section{Discussion}
\label{sec:dis}
We proposed a cascade framework for volumetric segmentation, called segmentation-by-detection which introduced an attention mechanism. The framework consists of a detection module for precise localization and a segmentation module. Different from other volumetric segmentation method which took an entire volume as input, our method exploited the attention model obtained from the detection module simultaneously. It allowed the network to distill information from the potential object region which increased the efficiency as well as reduced the disrupt of the surrounding noise. 
The detection module can be replaced by a human interactive tool where human provides precise region proposals. This interactive system can achieve even better performance in segmentation.
We will devote our effort on that in the future.

\section{Acknowledgement}
\label{sec:ack}
We would like to acknowledge NVidia corporation, that donated a Titan X GPU to our group enabling this research.


\bibliographystyle{IEEEbib}
\bibliography{strings,refs}

\end{document}